\documentclass[conference]{IEEEconf}

\input epsf
\usepackage{graphicx}
\usepackage{multirow}
\usepackage{titlesec}
\usepackage{balance} 
\usepackage{stfloats}
\usepackage{xcolor}

\renewcommand\thesection{\arabic{section}} % arabic numerals for the sections

\renewcommand\thesubsectiondis{\thesection.\arabic{subsection}}% arabic numerals for the subsections

\renewcommand\thesubsubsectiondis{\thesubsectiondis.\arabic{subsubsection}}% arabic numerals for the subsubsections

\usepackage{amsmath, amssymb}
\usepackage{stmaryrd}

\usepackage{amsthm}
\theoremstyle{definition}
\newtheorem{definition}{Definition}[section]

\newcommand{\timespace} { (t, x, y) }
\newcommand{\truefalse} { \{ \mathrm{true}, \mathrm{false} \} }
\newcommand{\dbracket}[1] { \llbracket #1 \rrbracket }
\newcommand{\interp}[1] { \mathcal{I}_{#1} }
\newcommand{\spec} { \mathrm{Spec.} }
\newcommand{\oddspec} { \mathrm{ODD~Spec.} }
\newcommand{\odd} { \mathrm{ODD} }
\newcommand{\lod} { \mathrm{LOD} }
\newcommand{\cod} { \mathrm{COD} }
\newcommand{\od} { \mathrm{OD} }

\begin{document}

\title{\textbf{\Large Formalization of Operational Domain and Operational Design Domain\\for Automated Vehicles\\}}

\author{Ali Shakeri\\
	\normalsize German Aerospace Center (DLR),\\
	Institute of Systems Engineering for Future Mobility,\\
	Escherweg 2, 26129 Oldenburg, Germany,\\
	\normalsize ali.shakeri@dlr.de\\
}
%+++++++++++++++++++++++++++++++++++++++++++

% make the title area
\maketitle

\begin{abstract}
	Specifying an Operational Design Domain (ODD) is crucial for safeguarding automated vehicle systems against conditions that exceed their capabilities. Yet, prior definitions of ODD have relied on ambiguous and unclear terms, resulting in numerous misunderstandings and misconceptions. This paper introduces a formal approach to clearly define the Operational Domain (OD) and ODD for automated vehicles. Furthermore, the absence of essential terms, such as the OD, has resulted in the creation of numerous terms that have made things more complicated and confusing. This level of complexity is unacceptable when it comes to developing safety-critical systems, where any uncertainty can lead to significant risks. This study addresses these deficiencies by providing a precise mathematical model of OD and clarifying its relationship with other terms. Also, by formalizing these terms, this work establishes a foundation for developing further concepts such as ODD specification and ODD monitoring, which are explained in this paper.
\end{abstract}

\IEEEoverridecommandlockouts
\vspace{1.5ex}

\IEEEpeerreviewmaketitle

\section{Introduction} \label{sec:introduction}
Automated systems are limited by their hardware and software, meaning they cannot operate in all environments or under all conditions. As a result, it is critical to define the safe operational domain for such systems and ensure they do not exceed their capabilities. This is where the concept of Operational Design Domain (ODD) becomes significant, safeguarding the systems against the broad operational domain.

Among various automated systems, defining ODD for Automated Driving Systems (ADS) is of greater importance due to their safety-critical nature. Additionally, as the level of autonomy \cite{saej3016} increases, the decision-making responsibility increasingly shifts from humans to machines. Therefore, ADS must be capable of handling more unexpected things that may go wrong in the real world. Consequently, such systems shall have an accurate and consistent model of their operational domain and ODD. Otherwise, any misinterpretation in the operational domain and ODD could lead to injuries or put lives at risk \cite{koopman2022safe}.

Due to its importance, the topic of ODD has been a subject of considerable interest among researchers and industries. The various studies and standards that have emerged from these efforts will be discussed in more detail in Section \ref{sec:relatedwork}. However, it is worth mentioning here that despite the efforts made so far, there is still confusion regarding fundamental concepts and their connection. This confusion leads to various issues, such as misinterpretation of terminology, development of new concepts based on misconceptions, and proliferation of terms and concepts in new research and standards. These issues can significantly delay development processes and increase costs and safety risks. In the following, we will describe each of these issues and their consequences.

SAE J3016 defines ODD as "operating conditions under which a given driving automation system or feature thereof is specifically designed to function, including, but not limited to, environmental, geographical, and time-of-day restrictions, and/or the requisite presence or absence of certain traffic or roadway characteristics" \cite{saej3016}. While the SAE's definition of ODD is a useful starting point, it lacks a clear definition for terms like "operating condition" and "restriction", making the definition ambiguous and opening the doors to misinterpretations. Such misinterpretation of terms could cause misunderstandings between different engineering teams, and even worse, the ambiguity could propagate to future standards. Consequently, this will increase not only costs and delays but also safety risks, which is unacceptable for safety-critical systems.

On the other hand, the lack of a clear definition of OD can lead to misconceptions. One common misconception is the use of the term "ODD taxonomy", while it is clear that ODD is a specific property of a vehicle system that alters with every alteration in sensor setup or system design decision. What the standardization committees, including the recent ISO 34503, refer to as the ODD taxonomy, is, in fact, a taxonomy for characterizing the operational domain attributes. This misconception may not seem significant, but it can actually affect future work. For instance, as of writing this paper, there is no known method for generating scenario descriptions based on operational domain attributes (the recent work from Zhang et al. only covers scenario generation based on ODDs \cite{Zhang2024ODD}).

Another problem with the SAE's definition is that it establishes no relationship between ODD and the OD. The lack of a definition for OD leads to the introduction of many other unnecessary terms with very vague definitions, such as "Operational World Model," "Operational Road Environment Models," and "subject vehicle models" in Czarnecki work \cite{czarnecki2018operational}. As another example, in an attempt to establish an ODD specification, Schwalb et al. \cite{schwalb2021two} encountered difficulties as they tried to formalize related concepts. The absence of a clear relationship between ODD and the operational domain led to the creation of new terms in their work, including "situation" and "facts", which themselves lacked proper definitions and therefore made the formalization hard to follow \cite{schwalb2021two}. Besides, this proliferation of terminology adds complexity and ambiguity and can hinder research and development efforts.

Furthermore, the three primary problems discussed earlier can have further negative impacts. Due to broken relations between existing concepts, it is not easy to develop other concepts. One important concept is ODD monitoring, which "is essential for the ADS to be able to decide on triggering the minimal risk manoeuvre (MRM) or issuing a transition demand by the ADS" \cite{iso2023}. Colwell et al. describe the purpose of ODD monitoring as "to determine whether or not the ADS is in a situation that it was designed to handle safely" \cite{colwell2018automated} while they do not clarify the meaning of "situation". Also, C. Sun et al. define ODD monitoring as "to monitor the vehicle states and driving environment that satisfy a specific ADS function requirement" \cite{sun2021acclimatizing}. However, it is unclear what "vehicle state" or "function requirement" means.

To address issues, this paper will use a formal method based on notable work by Olderog \cite{olderog2008real} to clarify concepts and their relations. This study begins by defining OD as a key concept. Then, we show how this concept simplifies the development of related concepts such as ODD specification and ODD monitoring. After describing the terms, we present a formal representation of them. This final step is necessary to clarify the relationship between these concepts with less ambiguity. Finally, this work demonstrates how the new formulation enables us to create a more accurate model of OD and ODD.

It is important to note that the current work strives to avoid creating a detailed and precise ODD specification or offering a description for operational domain taxonomy. For this reason, certain concepts have been intentionally simplified to effectively convey the main message, highlighting the importance of formal methods in this domain. The novelty of the current work is establishing clear relations between essential concepts in the field of ODD for automated vehicles. Furthermore, it enables us to describe other concepts using these main ones.

This study starts with an overview of related work in Section \ref{sec:relatedwork}. Then in Section \ref{sec:background} the fundamental terms are defined. Building upon this foundation, the concepts of ODD and ODD specification are also defined in Section \ref{sec:background}.
Next, the mathematical preliminaries are presented in Section \ref{sec:preliminaries}, essential for formally representing the defined concepts. After that, a formal representation of OD in Section \ref{sec:od} and ODD and ODD specification in Section \ref{sec:odd} is introduced.
Also, further concepts, such as ODD monitoring, are discussed in Section \ref{sec:odd}. Finally, concluding remarks and future work are provided in Section \ref{sec:conclusion}.

\section{Related Work} \label{sec:relatedwork}
The operational domain of autonomous vehicles (AVs) consists of many different dimensions. Researchers and organizations have proposed various taxonomies to categorize these dimensions and describe the operational domain. The National Highway Traffic Safety Administration (NHTSA) identified a preliminary set of attributes categorized into six main categories \cite{thorn2018framework}. Koopman and Fratrik emphasized the importance of defining the operational environment for AVs by listing critical aspects such as terrain characteristics and environmental conditions. They proposed a comprehensive list of factors relevant to describing the operational domain. Gyllenhammar et al. develop a framework to categorize and quantify the "operating conditions" \cite{gyllenhammar2020towards}. Other standardization committees have also attempted to define the operational domain for ADS by developing a taxonomy of its attributes \cite{avsc2020,pas1883,iso2023}.

The operational domain can be infinite because of the endless possibilities of traffic situations. Neurohr et al. suggest that an approach to scenario description is crucial for comprehending the operational domain of automated vehicles since it allows the organization and analysis of the infinite space \cite{neurohr2021criticality}. Moreover, the operational domain is not a static entity and can change over time. As a result, new classes, properties, and attributes may emerge. This has been discussed by Weshhofen et al., who suggest using ontologies to formally represent both the operational domain and critical phenomena in urban traffic scenarios \cite{westhofen2022using}.
In another research, Erz et al. suggest an ontology to bridge the ODD, scenario-based testing, and AV architecture \cite{erz2022towards}. By leveraging this ontology, the authors seek to provide systematic guidance for defining ODDs.

Efforts have been made to develop languages that define ODD for driving systems. Irvine et al. (2021) propose a structured natural language approach to defining ODD for Automated Driving Systems (ADS), aiming to enhance understandability and accessibility for a diverse range of stakeholders, including regulators and system designers. Schwalb et al. built upon Irvine's work, transitioning from a structured natural language format designed for clarity and accessibility towards a more formal representation aimed at programmatic execution. At the time of writing this article, the ASAM OpenODD standardization committee is actively working on developing a language to describe the ODD specification \cite{openodd2021}.

\section{Terminology and Definitions} \label{sec:background}
Historically, conventional vehicle systems, including cars, trucks, and motorcycles, have been designed to function within a wide range of infrastructures built on the earth's surface. In such vehicle systems, a human driver is responsible for handling external conditions such as environmental and dynamic traffic. With the advent of AVs, depending on their level of autonomy, they will take over some or all of these responsibilities. Therefore, an AV must have a model of the environment within which it operates, referred to as the operational domain in this work.

The automated vehicle will use sensors to perceive the operational domain and actuators (e.g., braking system) to interact with it. Consequently, the vehicle's operation is primarily limited by the quality of its sensors and the response of its actuators. Also, the qualities of OD, such as infrastructure, environmental conditions, and dynamic traffic, can significantly impact the AV's performance. Accordingly, engineers, infrastructure operators, and AVs must clearly understand the operational domain and its attributes. A formal model of the operational domain is necessary, but first, a definition of it is required.
\begin{definition}[operational domain] \label{def:OD}
	The operational domain for vehicle systems refers to the attributes of the physical surroundings in which the vehicles navigate, including the natural terrain and human-made infrastructure, environmental phenomena, and traffic conditions.
\end{definition}
According to the above definition, characterizing the operational domain attributes is crucial to better understanding it and creating an accurate model of it. This characterization is done through several standards that provide a taxonomy for operational domain attributes \cite{avsc2020,pas1883,iso2023}. However, it is worth mentioning that the ODD taxonomy standards, such as the recent ISO 34503, despite their nomenclature, actually offer a taxonomy for the operational domain.

Next, it is important to acknowledge that AV systems cannot function in all environments due to hardware or software constraints. An automated vehicle - excluding those classified as SAE level 5 \cite{saej3016} - is not designed to function throughout the entire operational domain. Rather, it can safely operate only within a specific and restricted region of the operational domain, known as the Operational Design Domain (ODD).
The ODD can be defined based on the operational domain, which is as follows:

\begin{definition} [Operational Design Domain] \label{def:ODD}
	An Operational Design Domain (ODD) for a specific vehicle system refers to a subset of the operational domain (see Definition \ref{def:OD}) within which the system is specifically designed and engineered to operate safely.
\end{definition}

The definition given above establishes a relationship between ODD and OD; that is, ODD for a particular system is a specific region within the OD.
For AV systems, it is essential to clearly and unambiguously specify this region for the system and the driver. Failure to do so could result in the vehicle encountering circumstances beyond its control or confusion in making a decision. For more detail, see completed investigation cases \cite{NTSB} that involved an AV crash.
An ODD specification is defined as follows:

\begin{definition} [ODD specification] \label{def:ODDSpec}
	An ODD specification for a specific system comprises a collection of declarative statements defined over OD attributes characterized by an OD taxonomy. These statements specify the ODD (see Definition \ref{def:ODD}) and ODD boundaries within OD.
\end{definition}

\begin{figure}
	\centering
	\includegraphics[width=\columnwidth]{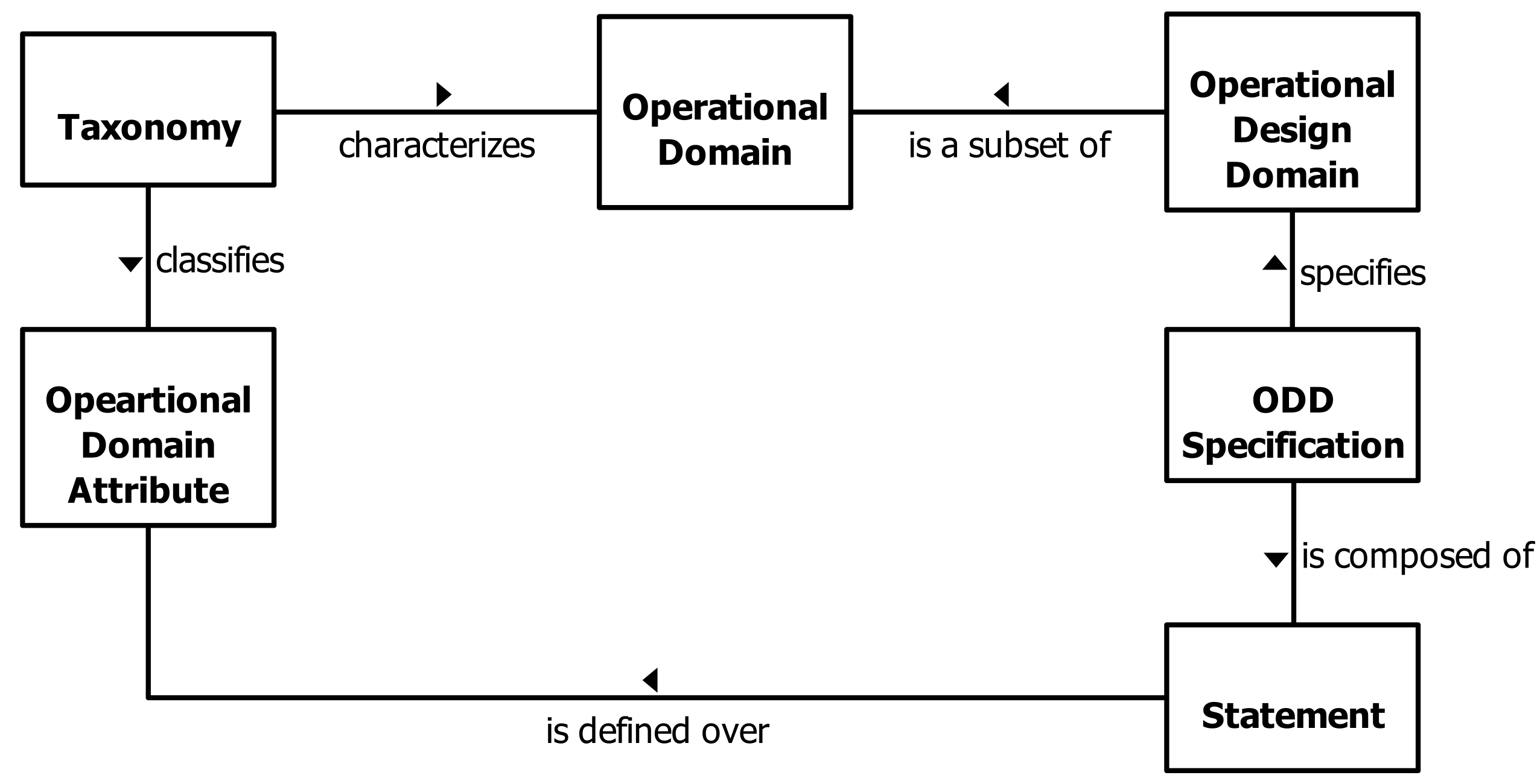}
	\caption{This illustration depicts the relationship between different concepts used in this work. The Operational Domain is characterized by a Taxonomy such as ISO 34503 \cite{iso2023}, which also classifies the Operational Domain attributes. An Operational Design Domain is a subset of the Operational Domain, specified by an ODD specification that is composed of a collection of Statements defined over attributes.}
	\label{fig:OD_ODD_relation}
\end{figure}

It is important to note that the terms "ODD" and "ODD specification" are often used interchangeably, but they actually have distinct meanings.
While ODD refers to the specific domain in which a system is intended to operate, ODD specification is a term used to define and specify that specific domain. Figure \ref{fig:OD_ODD_relation} provides a high-level illustration of the relationship between fundamental concepts used in the current study.

Furthermore, it is worth noting that an ODD could be an infinite set of values that is hard to represent in computer systems. Yet, ODD specification is a finite set of declarative statements, each declaring an acceptable range of values for OD attributes based on the system's limitation.

This study intentionally employs a simplified ODD example to focus on introducing the terms, concepts, and their mathematical representation. However, this will not undermine the applicability of the formalization method to practical cases. As an example, consider the following textual specification that specifies the ODD for a specific AV system using natural language:

\begin{align}
	&\textit{The system is designed and only allowed to operate} \nonumber \\ 
	&\qquad\qquad\textit {on motorways}, \tag{$s_1$} \label{eqn:s1} \\
	&\qquad\qquad\textit {where pedestrians are prohibited}, \tag{$s_2$} \label{eqn:s2} \\
	&\qquad\qquad\textit {up to speed of 60 km/h}. \tag{$s_3$} \label{eqn:s3}
\end{align}

The ODD specification, as described above, comprises three statements: \eqref{eqn:s1}, \eqref{eqn:s2}, and \eqref{eqn:s3} that presume the safe operation of the system only when it operates in motorways, in the absence of pedestrians, and with the maximum operational speed of $60\, \mathrm{km/h}$.

A natural language specification could lead to ambiguities and uncertainty when specifying the ODD. Therefore, the statements containing temporal and spatial constraints shall be expressed precisely. Accordingly, they shall be formalized using a formal specification method such as the one established by E.-R. Olderog and H. Dierks \cite{olderog2008real} for real-time systems. Besides, computer systems require a machine-readable specification. This study does not have the goal of creating a specification that can be read by machines. Instead, the ASAM OpenODD standardization committee \cite{openodd2021} is developing a language that can be used to describe OD and specify ODD. The remainder of this study focuses on formally representing OD, ODD, and ODD specifications.

\section{Preliminaries} \label{sec:preliminaries}
As discussed in previous section, formalization of concepts including OD, ODD, and ODD specification is necessary.
This formalization is based on some preliminary definitions which is provided in this section.
Here, the formal definition of attributes and statements is presented, which is the building block of other concepts.
\subsection{Attributes}
The operational domain can be described by a set of attributes denoted as $\mathbb{A}=\{A_1, A_2, \dots, A_n\}$ and their corresponding set of data types $\mathbb{D}=\{\mathcal{D}_1, \mathcal{D}_2, \dots, \mathcal{D}_n\}$.
Let $A$ be an attribute with data type $\mathcal{D}$, i.e., $A$ has a value in $\mathcal{D}$. For brevity, the notation $\mathcal{D} (A)$, is introduced to denote the data type of an attribute $A$.

To start with an example, consider ODD statements that were introduced in Section \ref{sec:background}.
The statement \eqref{eqn:s1}, implicitly assumes that among different "road types" the system is designed to operate on "motorway".
For this statement, we can abstract "road type" as an attribute and "motorway" as a value for this attribute.
For the sake of this example, let us assume that the attribute "road type", denoted by $A_1$, accepts three values, namely "motorway", "regional", and "rural".
In the same way, the statement \eqref{eqn:s2} can be described using the attribute "presence of pedestrian", denoted by $A_2$ that has a Boolean data type, i.e. it's value is "true" whenever there is a pedestrian on the road at a specific time and location and is "false" otherwise.
Finally, the statement \eqref{eqn:s3} can be described by an attribute "operational speed", denoted by $A_3$, that represents the speed the vehicle is allowed to reach.
This attribute is described with a real number data type.

The semantic of an attribute $A$, is given by an interpretation, $\mathcal{I}$, at a certain time, $t \in \mathrm{Time}$, and location, $(x, y) \in \mathrm{Space}$. 
The interpretation, $\mathcal{I}$, is a mapping that assigns to each attribute $A$, a value in $\mathcal{D}$,
\begin{equation}\label{eqn:interpretation}
	\mathcal{I} \colon \mathbb{A} \times \mathrm{Time} \times \mathrm{Space} \to \mathbb{D}.
\end{equation}
The value of $A$ for a specific interpretation $\mathcal{I}$ at time $t$, and location $(x, y)$, is denoted by $\mathcal{I}(A)\timespace$ or alternatively $\interp{A} \timespace$.
In relation \eqref{eqn:interpretation}, $\mathrm{Time}$ denotes the time domain which is a non-negative real number in $\mathbb{R}_{\geq 0}$, and $\mathrm{Space}$ denotes space domain as a tuple of two real numbers that are a subset of $\mathbb{R}^{2}$. For this a Geodetic system can be used such as WGS 84 that is being used in Global Positioning System (GPS) equipment \cite{slater1998wgs}.

\begin{figure*}[bp]
	\centering
	\includegraphics[width=0.7\textwidth]{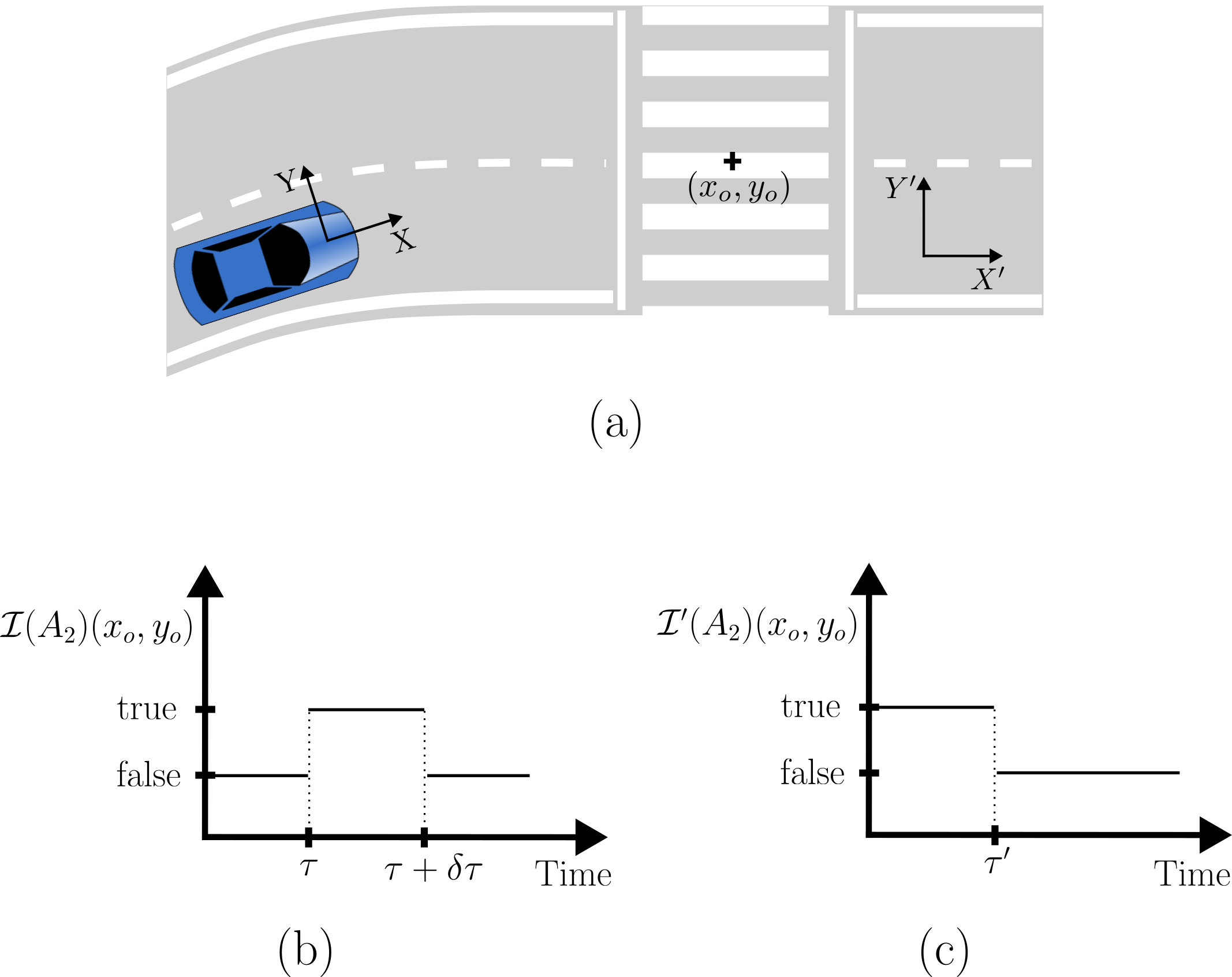}
	\caption{(a) shows a vehicle moving in a segment of a road while facing a pedestrian zone in front.
		A frame of reference $(X, Y)$ is attached to the vehicle and another frame of reference $(X^{\prime}, Y^{\prime})$ to the road.
		(b), (c) show two different interpretation of attribute $A_2$ representing the presence of a pedestrian at a certain location, labeled with $(x_o, y_o)$ coordinates denoted by a cross at sub-figure (a).}
	\label{fig:road}
\end{figure*}

In general, $\interp{A} \timespace \in \mathcal{D}(A)$, and by considering the example, the interpretation of attributes in statements \eqref{eqn:s1}, \eqref{eqn:s2}, and \eqref{eqn:s3} are respectively represented as
\begin{align}
	&\interp{A_1} \timespace \in \{ \mathrm{motorway}, \mathrm{regional}, \mathrm{rural} \}, \label{eqn:a1} \\
	&\interp{A_2} \timespace \in \truefalse, \label{eqn:a2} \\
	&\interp{A_3} \timespace \in \mathbb{R}. \label{eqn:a3}
\end{align}
Fig. \ref{fig:road} (b), (c) shows the fact that various interpretations of an attribute exists for a specific time and location.
It is worth noting that representing the location $(x, y)$, in space requires a fixed frame of reference.
Choosing a frame of reference is arbitrary and there is no preferred one.
Fig. \ref{fig:road} shows two frames of reference, one attached to the vehicle and another one attached to the road.

\subsection{Statements}
Statements are the building blocks of ODD specifications as they describe constraints on attributes.
An ODD specification is composed of set of statements $\mathbb{S}=\{S_1, S_2, \dots, S_n\}$.
Each statement $S \in \mathbb{S}$ is defined with the syntax, $\mathcal{A} \, \bowtie \, d $, where $\mathcal{A}$ is an attribute symbol, and $d$ belongs to data type $\mathcal{D}(A)$. Also, the $\bowtie$ symbol is one of the binary predicate symbols $\{=, <, >, \leq, \geq\}$, however, all binary predicate symbols might not be relevant for all attribute, or they may need to be defined. For instance, if the binary relation $>$ is not defined for the 'road type' attribute, then $A_1 > \mathrm{motorway}$ has no meaning.

A statement is defined with the following syntax
\begin{equation*}
	S ::= \mathcal{A} \bowtie d ~ | ~ \neg S,
\end{equation*}
therefore, if $S$ is an statement, then is $\neg S$.
%More complicated statements can be composed of logical conjunction of simpler statements, i.e., if $S_1$ and $S_2$ are statements, then is $(S_1 \land S_2)$.
For instance, the following relations have the syntax of a statement
\begin{align} 
	&S_1 := ~~ \big( A_1 = \mathrm{motorway} \label{eqn:f1} \big), \\
	&S_2 := ~~ \big( A_2 = \mathrm{false} \label{eqn:f2} \big), \\
	&S_3 := ~~ \big( A_3 < 60\, \mathrm{km\,h}^{-1} \big). \label{eqn:f3}
\end{align}
 
The semantics of a statement depend on the interpretation of its corresponding attribute. The semantics of statement $S$, denoted by $\mathcal{I}\dbracket{S}$, is a function returning the truth value of a statement, given an interpretation $\interp{A}$, that assigns a value to the corresponding attribute $A$ at certain time and space,
\begin{equation}\label{eqn:statementsemantic}
	\mathcal{I}\dbracket{\,} \colon \mathbb{S} \times \mathrm{Time} \times \mathrm{Space} \to \truefalse.
\end{equation}

For statements $S_1$, $S_2$, and $S_3$ defined in relations \eqref{eqn:f1}, \eqref{eqn:f2}, and, \eqref{eqn:f3}, $\mathcal{I} \dbracket{S_1}\timespace = \mathrm{true}$ whenever for $t \in \mathrm{Time}$, and, $(x, y) \in \mathrm{Space}$, an interpretation of $A_1$ is given such that $\mathcal{I}(A_1)\timespace = motorway$.
In the same way, $\mathcal{I} \dbracket{S_2}\timespace = \mathrm{true}$, whenever for a given interpretation, there is no pedestrian present on the road at $\timespace$, i.e., $\mathcal{I}(A_2)\timespace = \mathrm{false}$,
and $\mathcal{I} \dbracket{S_3}\timespace = \mathrm{true}$, whenever at $\timespace$ the speed of the vehicle is less than $60\, \mathrm{km/h}$.

\section{Operational Domain (OD)} \label{sec:od}
The operational domain of a system is fully realized when all its relevant attributes and their respective data types are specified. The process of identifying these relevant attributes is inherently dependent on the level of abstraction and the granularity of modeling details required for the system's operation. For instance, the specificity of attributes can vary widely based on standards and use cases; some standards might consider the 'type of asphalt' on which the vehicle operates as a relevant attribute due to its impact on vehicle handling and safety features. At the same time, others may deem this detail too granular and omit it from consideration.

Numerous standardization committees identified and classified Operational Domain (OD) attributes and their corresponding data types by providing a taxonomy of OD attributes. For instance, ISO 34503 \cite{iso2023} and BSI PAS 1883 \cite{pas1883} offered an OD taxonomy. However, it is important to highlight that they inaccurately named it an ODD taxonomy. The following section aims to remedy this confusion by formally representing an Operational Domain.

For an operational domain that is characterized by a set of attributes $\mathbb{A}=\{A_1, A_2, \dots, A_n\}$, it can be represented mathematically as a set of tuples over data types,
\begin{equation} 
	\label{eqn:od1}
	\od := \mathcal{D}_1 \times \mathcal{D}_2 \times \dots \times \mathcal{D}_n.
\end{equation}

For example if $A_1$ and $A_2$ are all relevant attributes of a certain system with range of possible values represented in Equations \eqref{eqn:a1} and \eqref{eqn:a2}, then OD of such system is represented by
\begin{align*} 
	\od = \{&\mathrm{(motorway, true), (motorway, false)}, \\
		    &\mathrm{(regional, true), (regional, false)}, \\
		    &\mathrm{(rural, true), (rural, false)} \}.
\end{align*}
The above example shows an OD defined over attributes with discrete data types.
In the same way, OD for attributes with continuous data types can be defined.
However, such OD forms an infinite set and an OD description is required to formally specify such an infinite OD.

A vehicle system explores different regions of the OD at various times and locations. In other words, the vehicle's sensors measure distinct values for OD attributes at multiple times and locations. To address this variability, it is useful to introduce a variable named Local Operational Domain (LOD), which is essentially an element of the OD. LOD is a tuple, given by interpretation of all measured attributes $\interp{A_1}, \interp{A_2}, \dots \interp{A_n}$ at a certain time $t$ and location $(x, y)$:
\begin{align}
	\label{eqn:lod}
	\lod \timespace :=  \big( \interp{A_1}, \interp{A_2}, \dots, \interp{A_n} \big).
\end{align}
For example, consider a simplified OD defined over $A_1$ and $A_2$. Let us assume that at time $t$ and location $(x, y)$, the type of road is motorway, and no pedestrian exists on the road. Then LOD would be represented as $\lod = (\mathrm{motorway, false})$.

Defining another quantity that is very similar to LOD but has a different meaning is beneficial. This quantity will be used later when introducing the concept of ODD monitoring. The Current Operational Domain (COD) can be defined as
\begin{equation}
	\cod := \lod ~(t_c, x_c, y_c)
\end{equation}
where $t_c$ is the current time and $(x_c, y_c)$ is the space coordinates at time $t_c$. The distinction between COD and LOD is that COD indicates the value of OD attributes in the current time and space, while LOD is a variable that can potentially indicate past or future of OD or value of OD at space coordinates that are different from current AV coordinates.

\section{Operational Design Domain (ODD)} \label{sec:odd}
After understanding essential concepts and mathematical preliminaries, this section is dedicated to the formal representation of ODD.
According to Definition \ref{def:OD} and Definition \ref{def:ODD}, it is clear that $\odd$ is a subset of $\od$.
However, a specification of ODD, denoted by $\spec$ is required according to Definition \ref{def:ODDSpec} to clearly specify ODD.

The formal specification of ODD denoted by $\spec$ is either a single statement (introduced in Section \ref{sec:preliminaries}) or logical conjunction of multiple statements or specifications defined by the following syntax: 
\begin{equation} \label{eqn:specsyntax}
	\spec ::= S ~ | ~ \neg \spec ~ | ~ \spec_1 \land \spec_2.
\end{equation}

Using the grammar in equation \eqref{eqn:specsyntax}, the statements \eqref{eqn:s1}, \eqref{eqn:s2}, and \eqref{eqn:s3} that are introduced in section \ref{sec:background}, can be described by \eqref{eqn:f1}, \eqref{eqn:f2}, and \eqref{eqn:f3} respectively, to form an ODD specification denoted as $\oddspec$:
\begin{equation} \label{eqn:ODDspec}
	\oddspec = S_1 \land S_2 \land S_3.
\end{equation}

Before moving forward, it is essential to note that an AV equipped with a set of sensors may only be able to measure some attributes of the OD. For instance, a vehicle might lack detectors for measuring smoke in the air. Additionally, a specific AV design might intentionally ignore certain OD attributes. For example, it may have a sensor setup that remains functional even in the presence of smoke. In the same way, an ODD specification can be ignorant of some OD attributes. 

In cases where ODD specifications have lower attribute coverage than OD attributes, additional assumptions must be made to prevent confusion when developing ODD specifications for such a system. One basic assumption that is called permissive assumption in the current study is that if the system is ignorant of particular attributes and their values, it implies that the system permits 'all' values for such attributes. For instance, if a permissive ODD specification for a specific system is silent about road types, it implies that the system can operate on all types of roads.

Other more elaborate assumptions could be made to restrict some or all attributes that are not mentioned in the ODD specifications; however, delving deeper into a specification language for ODD is beyond the scope of the current study. The interested reader is referred to the work by Schwalb et al. \cite{schwalb2021two} and Irvine et al. \cite{irvine2021two} for more details on specification languages for ODD specifications.

The semantic of an ODD specification, denoted by  $\mathcal{I}\dbracket{\spec}$ is a function that returns the truth value of a certain $\spec$, given the $\lod \in \od$ at a certain time and space for all relevant attributes:
\begin{equation}\label{eqn:semanticspec}
	\mathcal{I}\dbracket{\spec} \colon \od \times \mathrm{Time} \times \mathrm{Space} \to \truefalse.
\end{equation}
For brevity, $\mathcal{I} \dbracket{\spec} \big(\lod \timespace \big)$ is used to denote the function in Eq. \eqref{eqn:semanticspec}.

With the definition of Eq. \eqref{eqn:semanticspec}, ODD can be represented as all elements of OD that satisfy ODD specification as follows:
\begin{align} \label{eqn:ODD}
	\begin{split}
	\odd = \{ &\lod \timespace \in \od ~|~ \\
	&\exists ~\mathcal{I}, t, x, y \cdot \mathcal{I}  \dbracket{\spec}(\lod \timespace) \}.
	\end{split}
\end{align}
The procedure described above highlights the difficulty of representing ODD for a system without an appropriate definition and formal representation of OD and ODD specification.

\section{Discussion} \label{sec:discussion}
This study emphasizes the significance of formally representing ODD-related terms, particularly the operational domain (OD). As detailed in section \ref{sec:od}, a formal representation of ODD becomes feasible only after the mathematical representation of OD is established. This approach has proven essential in providing a clear and well-defined model, which effectively eliminates the ambiguities previously associated with these terms. This section will further explore the implications of our formalization approach, discussing how it helps to define other concepts and enhances the clarity.

Expanding upon the established foundation, it is possible to construct further advanced concepts. One of these concepts that is of great importance in the development and operation phases of AVs is the monitoring of the Operational Design Domain (ODD) or simply ODD monitoring. ODD monitoring is crucial because it ensures the AV operates safely within its specified operational boundaries. Continuous monitoring of the system is essential to prevent system failures or safety breaches. This means that the system must be monitored continuously to ensure that it is functioning within the ODD. Consequently, ODD Monitoring can be defined as follows:

\begin{definition}[ODD monitoring] \label{def:ODDmonitoring}
	ODD monitoring refers to the process of ensuring whether the current operational domain (COD) measured by an AV's sensors satisfies the ODD Specification defined for that specific AV.
\end{definition}

The relation \eqref{eqn:semanticspec} states that if $\mathcal{I} \dbracket{\spec}(\cod) = \mathrm{true}$, for a specific AV system, it implies that the system's current operational domain (COD) is within the boundaries defined with ODD specification.
Alternatively, according to relation \eqref{eqn:ODD}, it is possible to show that
\begin{equation*}
	\cod \in \odd \iff \mathcal{I} \dbracket{\spec}(\cod) = \mathrm{true}. 
\end{equation*}

In other words, ODD monitoring involves evaluating the truth value of ODD specifications in the current time and location.

The formalization approach presented here has other practical implications. This work illustrates how ODD specifications can be effectively related to OD attributes. Although the goal of current work was not to formulate a comprehensive ODD specification, it proposed a straightforward grammar of statements and ODD specifications. Additionally, it successfully demonstrated how these specifications can be evaluated with OD elements, as shown in Eq. \eqref{eqn:semanticspec}.

This study intentionally used a simplified approach to explain the importance of formal methods in safeguarding automated vehicles. However, this approach has limitations, and further investigation and extension are needed.

Firstly, more detailed statements and specifications can be added while maintaining the same formal approach. For instance, conditional or time-dependent statements can be introduced specifically tailored to address various real-world scenarios automated vehicles might encounter.

Second, the measurement of the operational domain and current operational domain (COD) demands a thorough investigation due to the numerous unanswered questions concerning measurement techniques, which remain largely unstandardized. For instance, the method for accurately measuring the position of an object on the road is crucial and requires detailed consideration. Should such objects be regarded merely as points, or is it more practical to consider an effective radius that better reflects their physical presence? Such methods shall address the measurement error and provide ways to safeguard AV operations against these measurement errors.

Finally, building upon the current formalization framework, it is possible to define additional concepts, such as degraded functionalities and restricted ODD regions. By identifying specific subsets of the ODD that apply to degraded operation modes, as suggested in the literature \cite{colwell2018automated}, we can tailor the system's responses to various levels of functionality impairment. Exploring these ideas in future research requires the formalization approach presented in this paper.

\section{Conclusion} \label{sec:conclusion}
This initial study has provided a basis for understanding the term OD and its importance in clarifying related terms such as ODD and ODD specification. By introducing precise definitions and a structured approach, we have addressed the ambiguities that previously clouded these critical terms, enhancing the clarity crucial in developing automated vehicle systems. In this regard, this work introduced a preliminary formal representation of OD and ODD and explained these notions using several examples. In addition, the current study demonstrated the procedure for creating a primary ODD specification by employing basic statements and evaluating it based on the operational domain. In the end, it shows how other concepts, such as ODD monitoring, can be built on top of the current formalization.

Despite the progress made, this study acknowledges the limitations of the current formalization and the need for extensions. In particular, there is a need to create a comprehensive language for ODD specification that addresses technical intricacies, such as the inclusion of conditional statements, which were beyond the scope of this work. Also, the temporal aspects of ODD statements (such as time intervals during which some statement is not fulfilled) need to be explored in future work. Nonetheless, future work will benefit the approach presented here, incorporating more detailed and dynamic specifications and exploring measurement techniques for real-world applications.

\section{Acknowledgment}
This work received support from the GAIA-X 4 PLC-AAD project, which is funded by the German Federal Ministry of Economic Affairs and Climate Action (BMWK) under the funding code 19S21006K. The author is grateful to Dr. Bernd Westphal for his insightful and valuable discussions and guidance throughout the research process. Additionally, appreciation is due to Anna Austel, Ishan Saxena, and Dominik Grundt for their critical feedback on earlier drafts of this paper. Their perspectives and critiques were essential in refining the arguments and enhancing the overall quality of the manuscript.

\bibliographystyle{IEEEtran}
\bibliography{references}

\balance

\end{document}